\documentclass[doublecolumn]{IEEEtran}

\usepackage{graphicx}
\usepackage[percent]{overpic}%added by Zhengtong
\usepackage{subfig}
\usepackage{amsmath}
\usepackage{amsthm} % definte the proof environment: block letters + italic letters
\usepackage{amssymb}
\usepackage[font=footnotesize]{caption} % set the captain font size to 8 (i.e. footnotesize)
\usepackage{subfig} % uses subfloats within a single float MUST after the package {caption}!!
\usepackage[noadjust]{cite} % alphabetize grouped citations automatically [3, 1, 2] to [1, 2, 3]
\usepackage{color}
\usepackage{algorithm} % options: boxed [section]
\usepackage{algpseudocode} % for algorithm
\usepackage{enumerate}
\usepackage[hidelinks,colorlinks=false]{hyperref}
\usepackage[left=0.75in, right=0.75in, top=0.75in, bottom=0.75in]{geometry}
\usepackage{authblk} % for authors' affiliation \affil{xxx}
\usepackage{arydshln} % for dashline in table or matrix

\usepackage{bm}
\usepackage{tikz}
\usetikzlibrary{calc} % for calculation functions in Tikz let, in commands in Tikz
\usetikzlibrary{shapes} % for block diagram
\usetikzlibrary{chains}
\usetikzlibrary{fit}
\usetikzlibrary{arrows}
\usetikzlibrary{decorations.text} % text along path
\usetikzlibrary{decorations.markings}
\usetikzlibrary{decorations.pathmorphing} % for zigzag arrow
\usetikzlibrary{shadows}
\usetikzlibrary{patterns}
\usetikzlibrary{matrix}
\usepackage{pgfplots}
\usepackage[europeanresistors]{circuitikz}
\usepackage[outline]{contour} % white surround text
\contourlength{1.5pt}

\graphicspath{{figures/}}

\usepackage{graphicx}
\usepackage{caption}
\usepackage{verbatim}
\usepackage{multirow}
\usepackage{subcaption}

\begin{document}
\title{{UniT: Data Efficient Tactile Representation with Generalization to Unseen Objects}}
\author{Zhengtong Xu, Raghava Uppuluri, Xinwei Zhang, Cael Fitch, Philip Glen Crandall, Wan Shou, Dongyi Wang, Yu She$^{*}$
\thanks{$^{*}$Address all correspondence to this author.}
\thanks{Zhengtong, Raghava, Xinwei, Cael, and Yu are with Purdue University, West Lafayette, USA  (E-mail: \{xu1703, ruppulur, zhan4645, fitch2, shey\}@purdue.edu).}
\thanks{Philip, Wan, and Dongyi are with University of Arkansas, Fayetteville, USA  (E-mail: crandal, wshou, dongyiw@uark.edu).}
\thanks{{This work was partially supported by the United States Department of Agriculture (USDA; No. 2023-67021-39072, 2023-67022-39074, 2023-67022-39075, and 2024-67021-42878) as well as National Science Foundation (NSF; No. 2423068). This article solely reflects the opinions and conclusions of its authors and not of USDA and NSF.}}}

\maketitle

\begin{abstract}
UniT is an approach to tactile representation learning, using VQGAN to learn a compact latent space and serve as the tactile representation. It uses tactile images obtained from a single simple object to train the representation with generalizability. This tactile representation can be zero-shot transferred to various downstream tasks, including perception tasks and manipulation policy learning. Our benchmarkings on in-hand {3D pose and 6D pose estimation tasks and a tactile classification task} show that UniT outperforms existing visual and tactile representation learning methods. Additionally, UniT's effectiveness in policy learning is demonstrated across three real-world tasks involving diverse manipulated objects and complex robot-object-environment interactions. Through extensive experimentation, UniT is shown to be a simple-to-train, plug-and-play, yet widely effective method for tactile representation learning. For more details, please refer to our open-source repository {https://github.com/ZhengtongXu/UniT} and the project website {https://zhengtongxu.github.io/unit-website/}.

\end{abstract}
\begin{IEEEkeywords}
Representation learning, tactile sensing, imitation learning.
\end{IEEEkeywords}

%%%%%%%%%%%%%%%%%%%%%%%%%%%%%%%%%%%%%%%%%%%%%%%%%%%%%%%%%%%%%%%%%%%%%%%%%%%%%%%%
\section{Introduction}\label{sec:intro}

Imitation learning has shown promising results equipping robots in the domain of manipulation, with the ability to master complex, high-precision, dexterous skills through human demonstrations \cite{chi2023diffusionpolicy,zhao2023learning}. However, the predominant focus of current research is on leveraging image or point cloud inputs to perceive scene information. During manipulation, robots encounter varied force interactions from both the environments and the manipulated objects. The sole dependence on images \cite{chi2023diffusionpolicy,zhao2023learning} or point clouds \cite{ze20243d} could obscure critical details about the objects' in-hand states and the dynamics of force interactions, which may be crucial for effective manipulation. Despite the current focus on visual information in robot learning research, humans routinely use visual and tactile feedback in manipulation tasks. Therefore, exploring the integration of visual and tactile modalities in imitation learning could potentially enhance the robots' performance in manipulation tasks.

Vision-based tactile sensors like GelSight \cite{yuan2017gelsight} provide high-resolution feedback, capturing rich tactile information. Prior research suggests that GelSight’s unique sensing principles enable trained neural networks to generalize effectively. For instance, a gradient estimation model trained solely on images of a small ball contacting the sensor generalizes across objects with various shapes and textures \cite{wang2021gelsight}. Likewise, a tactile-reactive grasping policy trained on standardized texture-less blocks extends to diverse objects \cite{xu2024letac}. While these studies highlight the distinctiveness of tactile images compared to standard visual images, their generalizability remains task-specific. This naturally raises the following question: {can we use a single simple object to learn a data efficient tactile representation of one type of GelSight that 1) possesses generalizability, 2) incorporates as much of the rich information present in tactile images as possible, and 3) can be applied in a zero-shot manner across a variety of downstream tasks involving different objects?}

{To address this question, we propose UniT, data efficient tactile representation with generalization to unseen objects.} Our contributions are summarized as follows:

1. \textbf{Easy to Train yet Broadly Applicable}: We introduce UniT, a tactile representation learning method for GelSight sensors. UniT uses VQGAN to learn a structured latent space and serve as the tactile representation. It can utilize data on a single simple object to learn a data efficient tactile representation that can be generalized to objects of different sizes and shapes. Through experiments in image reconstruction involving diverse objects, we demonstrate that the representation learned by UniT from a single simple object can capture information of unseen objects on contact configurations, object shapes, and dynamic marker motions induced by applied forces. This kind of generalizability makes UniT both straightforward to train and broadly applicable.

2. \textbf{Deploy for Tactile Perception}: The tactile encoder trained through UniT can be seamlessly transferred to downstream tactile perception tasks. Furthermore, due to UniT's lenient data requirements during training, representations learned even from the simplest objects, such as a small ball, can effectively facilitate tactile perception of everyday objects. In experiments involving the task of estimating the in-hand 6D pose of the USB plug, UniT outperforms training of a ResNet \cite{he2016deep} from scratch, existing visual representation learning methods BYOL \cite{grill2020bootstrap} and MAE \cite{he2022masked}, and the state-of-the-art tactile representation learning method, T3 \cite{zhao2024transferable}.

3. \textbf{Effective Visual-Tactile Policy Learning}: In our framework, UniT can be integrated into visual-tactile imitation learning pipelines, achieving high-precision manipulation tasks with rich interactions. The experimental results show that for tasks involving substantial robot-object-environment interactions, policies incorporating UniT outperform those based solely on vision and those that treat tactile images as regular visual inputs.
\section{Related Work}

\subsection{Tactile-involved Imitation Learning}

Research that incorporates tactile arrays \cite{guzey2023see,lin2024learning} and force-torque feedback \cite{yang2023seq2seq} into imitation learning policies have shown strong performance in tasks with complex interactions between environments, objects, and robots. However, tactile interactions are inherently complex and these modalities often do not capture the full spectrum of tactile information. For example, tactile arrays may struggle to accurately perceive an object’s geometry, in-hand position, and orientation, while force-torque may not effectively capture dynamic force distributions on the hand/finger.

Additionally, vision-based tactile sensors such as GelSight \cite{yuan2017gelsight} offer high-resolution feedback that captures extensive tactile information. Recent studies incorporating GelSight into imitation learning have demonstrated its considerable potential and enhanced performance \cite{wang2024poco,yu2023mimictouch}.

\begin{figure}[t]
\centering
\begin{overpic}[trim=0 70 0 0,clip, width=0.5\textwidth]{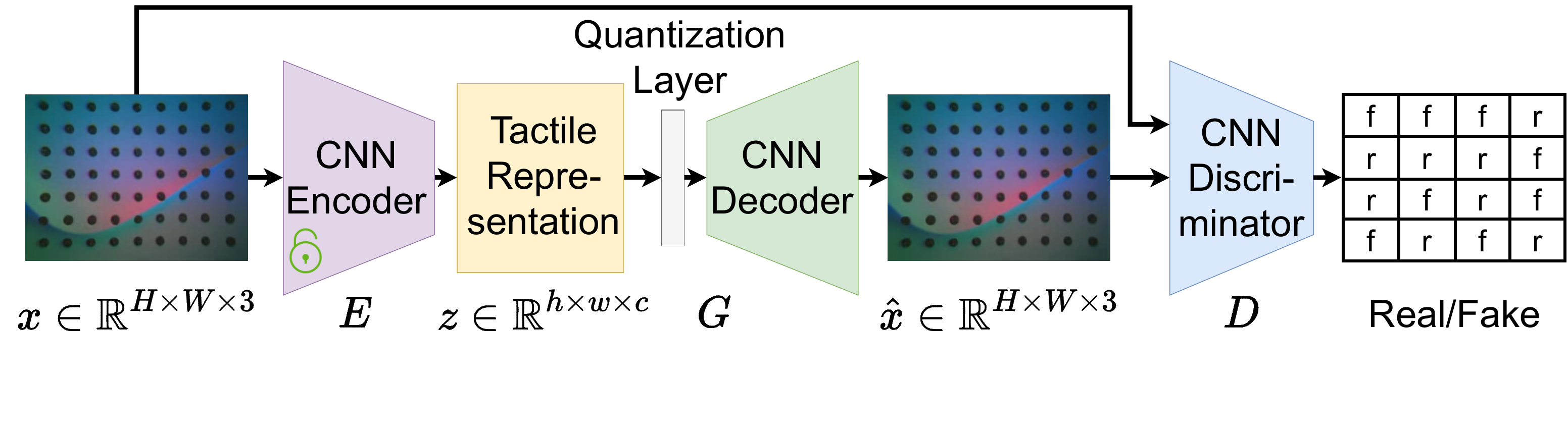}
\end{overpic}
\caption{Pipeline of UniT representation training.}
\label{fig:vqgan}
\end{figure}

\subsection{Representation Learning in Imitation Learning and Tactile Sensing}

Recent studies demonstrate that representation learning significantly enhances performance in imitation learning \cite{li2023crossway, pari2021surprising}. Encoders that are pretrained with representation learning are particularly adept at extracting information from visual observations.

In the field of tactile sensing, a variety of representation learning frameworks have been introduced, including the use of MAE \cite{cao2023learn, sferrazza2023power} and CNN \cite{polic2019convolutional}. Furthermore, the works in \cite{zhao2024transferable,higuera2024sparsh} introduce frameworks for tactile representation learning that scales across multi-sensors and multi-tasks.
The work in \cite{yang2024unitouch} learns multimodal tactile representation by aligning tactile embeddings to pretrained image embeddings associated with a variety of other modalities.

In this paper, UniT demonstrates that tactile images from a single, simple object can train a highly generalizable tactile representation—a capability absent in existing work.

\section{{Background: VQGAN autoencoder}}

{The work in \cite{esser2021taming} introduces VQGAN for high-resolution image synthesis. VQGAN integrates a VQVAE \cite{van2017neural} with a patch-based discriminator. Subsequently, the work in \cite{rombach2022high} introduces latent diffusion model, which leverages the VQGAN autoencoder to compress images into a latent space where diffusion and denoising processes can occur. The efficacy of the latent diffusion model is largely due to the robust image compression capabilities of VQGAN, which not only facilitates high-quality image reconstruction but also avoids high degree of variance in the learned latent space through VQ regularization.}

{Although VQGAN demonstrates impressive performance in generative model and image generation areas, its role as an autoencoder is primarily to enable high-quality image compression and reconstruction. In this paper, we demonstrate that VQGAN can serve as a highly effective tactile representation learner that can be trained with minimal data.}

\section{Method}

\begin{figure}[t]
\centering
\begin{overpic}[trim=0 0 0 0,clip, width=0.45\textwidth]{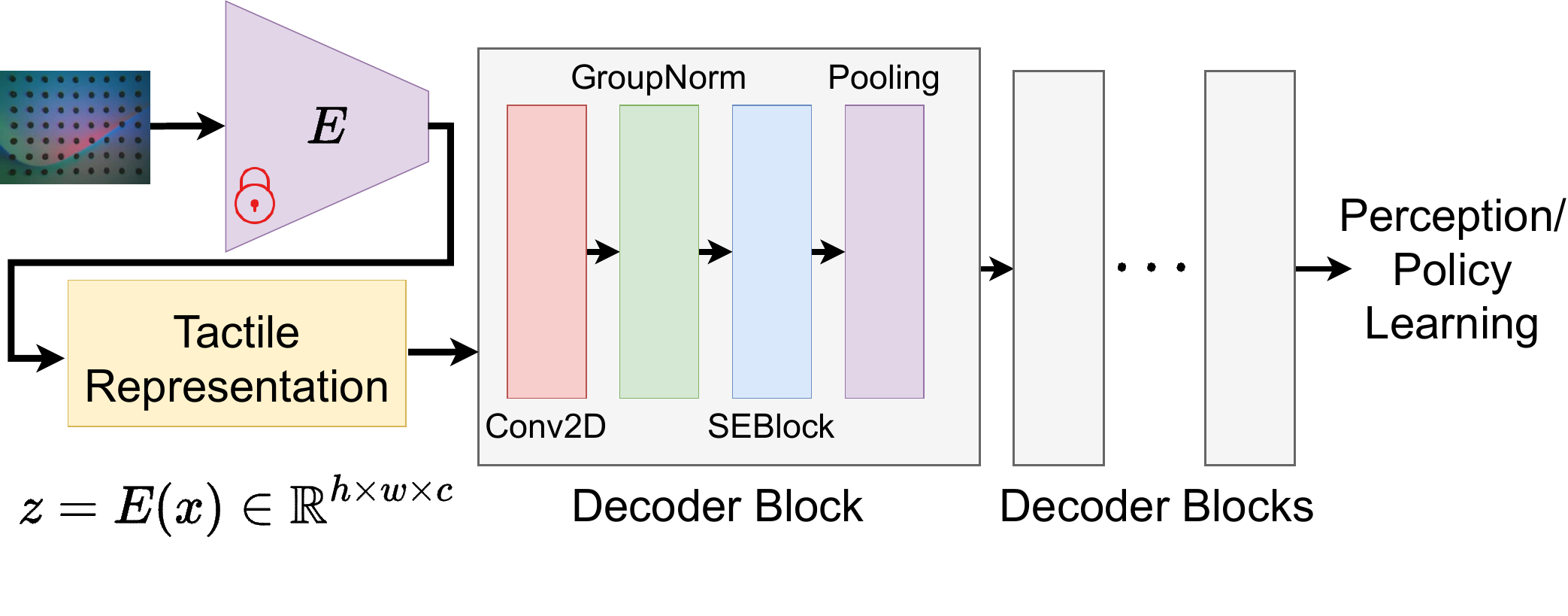}
\end{overpic}
\caption{Decoder architecture of implementing UniT representation to downstream tasks.}
\label{fig:decoder}
\end{figure}

\begin{figure*}[t]
\centering
\begin{overpic}[trim=0 0 0 0,clip, width=0.88\textwidth]{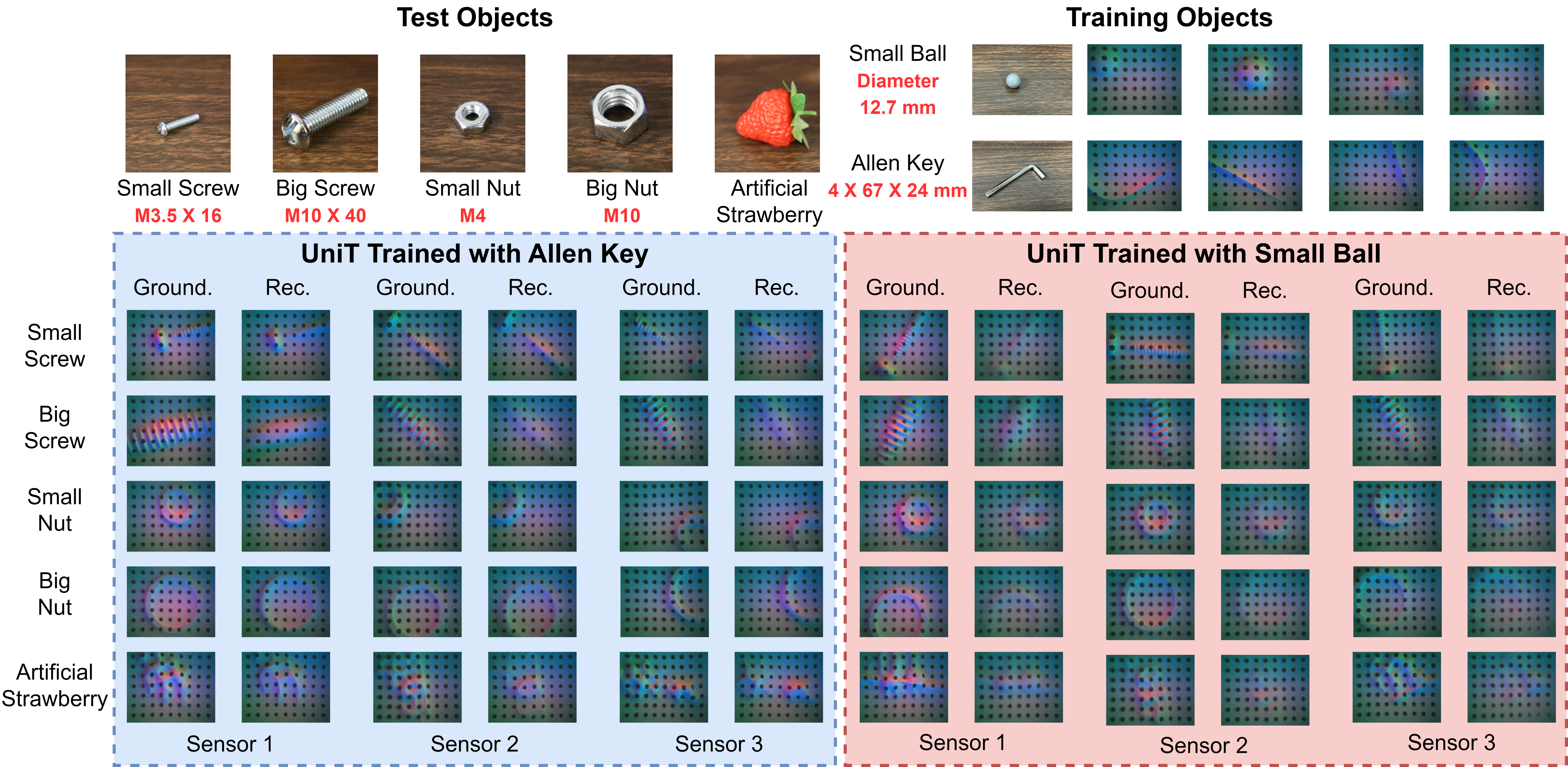}
\end{overpic}
\caption{Example results of UniT reconstruction of diverse unseen objects. Rec. represents reconstruction while Ground. represents ground truth. Sensor 1, 2, and 3 are three different GelSight minis. One training dataset for the autoencoder is only collected on one sensor.}
\label{fig:rec_unit}

\begin{overpic}[trim=0 0 0 0,clip, width=0.84\textwidth]{images/rec_mae.pdf}
\end{overpic}
\caption{Example results of MAE reconstruction of diverse unseen objects. We show the results of MAE with a ViT-Base backbone and a mask ratio 0.75 which is suggested by \cite{he2022masked}.}
\label{fig:rec_mae}
\end{figure*}

\begin{figure}[t]
\centering
\begin{overpic}[trim=0 0 0 0,clip, width=0.5\textwidth]{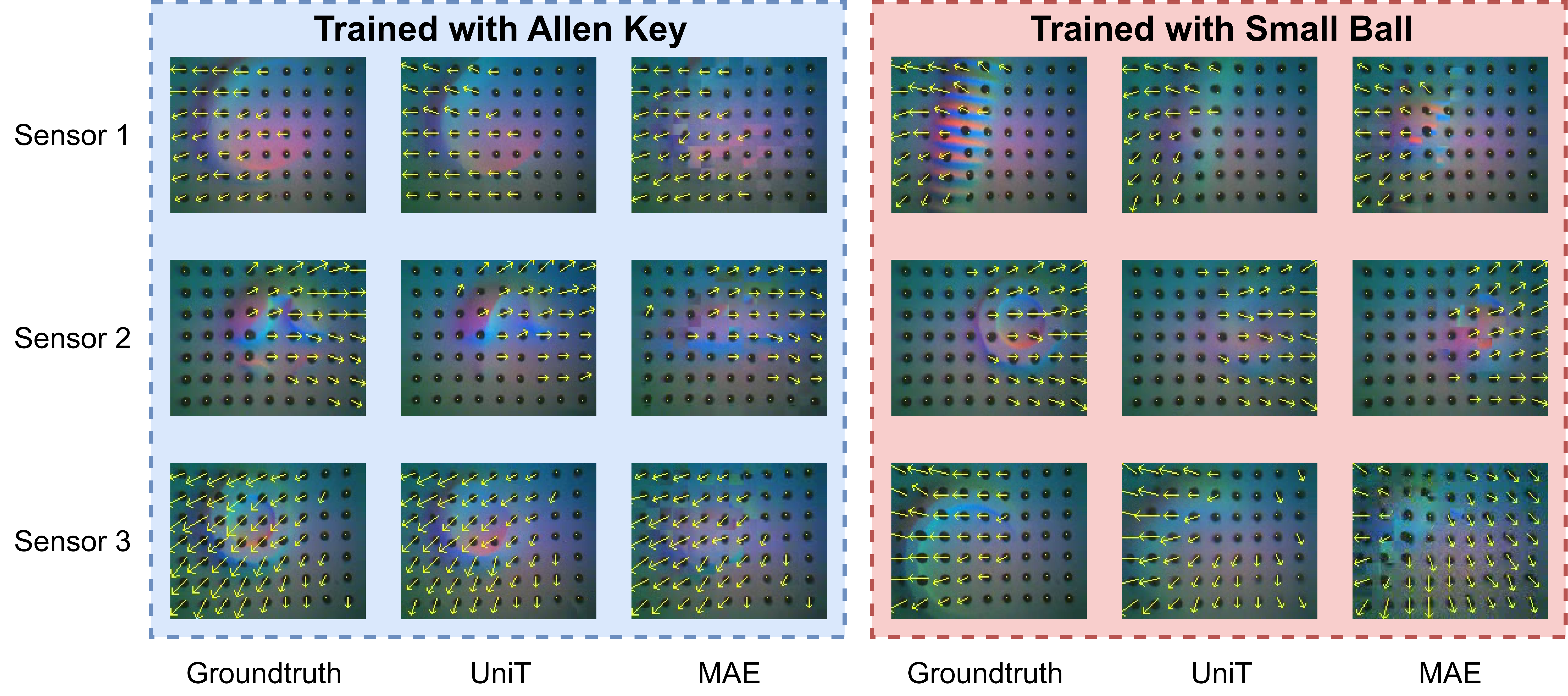}
\end{overpic}
\caption{Example results of marker tracking. To evaluate if the learned representations consist of information of the dynamic marker motion, we implement marker tracking \cite{yuan2017gelsight} on ground truth images and the corresponding image reconstructions by MAE and UniT.}
\label{fig:markers}
\end{figure}

\begin{figure*}[t]
\centering
\begin{overpic}[trim=0 0 0 0,clip, width=0.9\textwidth]{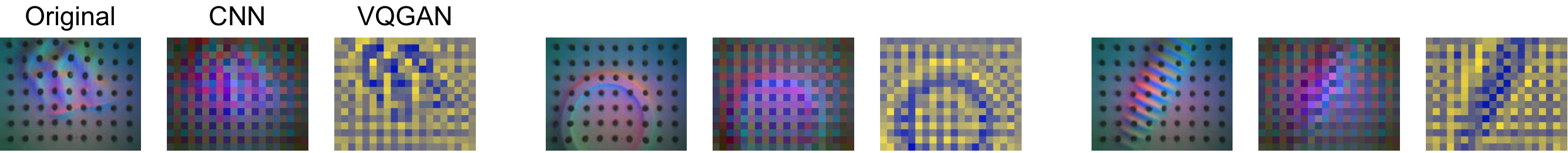}
\end{overpic}
\caption{{Visualization of latent spaces. We transformed tactile images of size 3×128×160 into latent spaces of size 3×16×20 and visualized these latent spaces as RGB images. Both VQGAN and CNN autoencoders are trained solely on the Allen key dataset. We tested three different unseen objects. The three images on the left are from artificial strawberry, the three in the middle are from nut, and the three on the right are from screw.}}
\label{fig:vis}
\end{figure*}

\subsection{Training Pipeline}

{Tactile images are created based on a unique imaging principle that combines RGB-tricolor light scattering with gel deformation. Unlike visual images, however, tactile images exhibit a much more compact color distribution. This compactness arises because the imaging process disregards object and background colors, focusing solely on tactile features such as contact configuration, geometry, and force distribution. Given the inherently low-variance color distribution in raw tactile images, we believe that VQGAN can effectively learn a mapping from raw tactile images to a latent space, and that this mapping can be achieved with a small amount of tactile data. Moreover, since VQGAN is designed to learn a low-variance latent space, we propose that for tactile images, such a latent space is not only sufficient to represent the raw tactile images but also enhances data efficiency and applicability for downstream tasks due to its structured characteristics. Accordingly, we leverage VQGAN for learning tactile representations.}

As depicted in Fig.~\ref{fig:vqgan}, the representation learning pipeline employs a VQGAN architecture with the quantization layer absorbed into the decoder. This pipeline comprises a CNN encoder ${E}$, a CNN decoder incorporating the quantization layer ${G}$, and a patch-based discriminator ${D}$. The entire framework is trained in a self-supervised manner. Denote the tactile image input as $x \in \mathbb{R}^{{H} \times {W}\times {3}}$, where the encoder ${E}$ maps the image to a tactile representation $z = {E}(x) \in \mathbb{R}^{{h} \times {w}\times {c}}$. The decoder reconstructs $z$ to an image $\hat{x}$.

The use of a discriminator is justified because recent studies \cite{fei2023masked} have shown that adversarial loss can enhance the performance of representation learning.  The presence of VQ regularization enables the learning of a structured tactile representation with reduced variance, which can improve the performance of representation learning based on the experimental results in Section~\ref{sec:perception_exp}.

\subsection{Train with Simple and Single Object}\label{sec:train_simple}

In this section, training strategy is outlined as: training the autoencoder depicted in Fig.~\ref{fig:vqgan} using a single simple object to significantly simplify data collection. However, the UniT representation obtained through this training method exhibits generalizability. 

In this paper, GelSight mini with markers is used. We believe that other sensors from the ``GelSight family" can also use our framework, as they share the same sensing principles. We chose the sensor gel with markers because it improves the tactile image's capacity to capture the distribution of shear forces.

We present two training examples here: datasets were collected for two types of objects, an Allen key and a small ball, and autoencoders were trained for each, respectively. As depicted in Fig.~\ref{fig:rec_unit}, both the Allen key and the small ball have simple shapes and lack surface texture, suggesting that other similarly rigid objects could also be suitable for training. This leniency in selecting training objects demonstrates the simplicity and straightforwardness of our data acquisition and training processes.

The specific data collection process involves capturing images from the GelSight sensor at a fixed frequency 10 Hz, contacting the object with the GelSight, and continuously changing the contact configuration and the magnitude of the applied force. 

The final size of our collected datasets includes 10,854 images of the Allen key and 4,831 images of the ball. Note that recording images at a frequency of 10 Hz enables the acquisition of such a dataset for a single object for no more than 20 minutes.

\subsection{Decoder Head for Downstream Tasks}

After completion of the autoencoder training, both the decoder and the discriminator are discarded. The encoder is subsequently connected to the proposed decoder blocks and then to downstream tasks like perception and policy learning tasks.

As shown in Fig.~\ref{fig:decoder}, for decoding the representation $z \in \mathbb{R}^{h \times w \times c}$, we utilize a decoder block consisting of \texttt{Conv2D}, \texttt{GroupNorm} \cite{wu2018group}, \texttt{SEBlock} \cite{hu2018squeeze} as the basic module.

In the experimental section of this paper, we demonstrate that after the representation learning phase is complete, freezing the UniT encoder and training only the decoder head for downstream tasks achieves performance comparable to fine-tuning the entire model.

\section{{Reconstruction Experiments}}
After completing the training for representation learning, experiments were conducted on image reconstruction using trained autoencoders across a series of unseen objects, as shown in Fig.~\ref{fig:rec_unit}. Moreover, comparing the performance of UniT with other methods, we conducted the same experiments using MAE \cite{he2022masked}, as illustrated in Fig.~\ref{fig:rec_mae}.

Our experimental results indicate that although UniT was trained only on a single simple object, the learned tactile representation can effectively generalize to unseen objects with diverse shapes, sizes, and textures. This tactile representation can reconstruct images that preserve most of the critical information of the original image, such as contact geometry and configuration, as shown in Fig.~\ref{fig:rec_unit}. Moreover, the representation trained with an Allen key performs better than the one with the small ball. This is logical, as a small ball is one of the simplest objects from which to extract features: it lacks texture, edges, distinct shapes, and is omnidirectionally symmetrical. Despite this simplicity, the representation trained using a small ball can still capture features of unseen objects to a certain degree. 

{As shown in Fig.~\ref{fig:rec_unit}, the representation learned by UniT can generalize across different units of the same type of sensor for image reconstruction, even when the representation is trained on a single object with one sensor unit.} {UniT does not support transfer across different sensor types with drastically different image characteristics, as it relies on minimal data from a single sensor type. However, UniT’s data collection and training are efficient, requiring only a small dataset from one object on a single sensor. Once trained, the representation can be applied to diverse downstream tasks of this type of sensor. More results of downstream tasks will be shown in Sections \ref{sec:perception_exp} and \ref{sec:policy_exp}.}

Compared to UniT, MAE performs less ideal. In some cases, it can reconstruct the rough shape and orientation of the original object, but in other instances it fails to reconstruct. For example, as shown in Fig.~\ref{fig:rec_mae}, some artificial strawberry images are reconstructed as small ball images.

Finally, the marker tracking results are shown in Fig.~\ref{fig:markers}, that demonstrate both UniT and MAE effectively capture the dynamic motion of markers. This capability is essential for applying these tactile representations in robot manipulation tasks with rich force interactions.

 {We also conduct an experiment to visualize the latent spaces, providing an illustration of the working mechanism of our proposed tactile representation, as shown in Fig.~\ref{fig:vis}. Specifically, we train a pure CNN-based autoencoder and a VQGAN autoencoder on Allen key data and use their respective encoders to project the same original image into different latent spaces. Specifically, we transformed tactile images of size 3×128×160 into latent spaces of size 3×16×20 and visualized these latent spaces as RGB images. The results reveal key differences: without VQ, the latent space retains a color distribution highly similar to the original image. This outcome is expected, as reconstruction training without VQ relies on a pure CNN-based autoencoder, where convolutions inherently preserve the original color distribution. In contrast, the VQGAN-based latent space exhibits a significantly more structured representation. While preserving essential characteristics of the original image, VQGAN amplifies meaningful features while suppressing background color variations, resulting in a more distinct, dichotomous visualization. Unlike the CNN-based latent space, which retains background color variance, the VQGAN-encoded representation focuses primarily on salient features, aligning with its ability to learn a low-variance latent representation.}

\section{Tactile Perception Experiments}\label{sec:perception_exp}

\begin{table*}[t]
\centering
\begin{minipage}{\textwidth}
\centering
\begin{tabular}{|c||c||ccc||ccc||cc||cc||cc||}
\hline
\multirow{3}{*}{} & \multirow{3}{*}{BYOL} & \multicolumn{3}{c||}{MAE-ViT-Tiny} & \multicolumn{3}{c||}{MAE-ViT-Base}  & \multicolumn{2}{c||}{UniT w/o VQ} & \multicolumn{2}{c||}{{UniT w/o Dis.}} & \multicolumn{2}{c||}{UniT} \\ \cline{3-14} 
                  && \multicolumn{3}{c||}{Mask Ratio} & \multicolumn{3}{c||}{Mask Ratio} & \multicolumn{2}{c||}{{Rep. Dim.}} &\multicolumn{2}{c||}{{Rep. Dim.}} & \multicolumn{2}{c||}{{Rep. Dim.}} \\ \cline{3-14}
                  && \multicolumn{1}{c|}{0.25} & \multicolumn{1}{c|}{0.5} & 0.75 & \multicolumn{1}{c|}{0.25} & \multicolumn{1}{c|}{0.5} & 0.75 & \multicolumn{1}{c|}{$8 \times 10$} & $16 \times 20$ & \multicolumn{1}{c|}{{$8 \times 10$}} & {$16 \times 20$} & \multicolumn{1}{c|}{$8 \times 10$} & $16 \times 20$\\ \hline
{A. K.}         &1.33& \multicolumn{1}{c|}{0.337} & \multicolumn{1}{c|}{0.358} & 0.374 & \multicolumn{1}{c|}{0.247} & \multicolumn{1}{c|}{0.273} & 0.370 & \multicolumn{1}{c|}{0.225} & 0.189 & \multicolumn{1}{c|}{{0.190}} & {0.145}& \multicolumn{1}{c|}{0.156} & \textbf{0.128}  \\
{S. B.}      &1.06& \multicolumn{1}{c|}{0.435} & \multicolumn{1}{c|}{0.685} & 0.789 & \multicolumn{1}{c|}{0.526} & \multicolumn{1}{c|}{0.560} & 0.388 & \multicolumn{1}{c|}{0.283} & 0.185 & \multicolumn{1}{c|}{{0.261}} & {0.193}& \multicolumn{1}{c|}{0.274} & \textbf{0.166}  \\ \hline
\end{tabular}
\end{minipage}

\vspace{0.5cm} 

\begin{minipage}{\textwidth}
\centering
\begin{tabular}{|l|l|l|l|l|l|l|l|}
\hline
ResNet34 w/o pretrain & ResNet34 with pretrain & T3 Tiny& T3 Small& T3 Medium & T3 Large & UniT SmallBall & UniT AllenKey \\ \hline
0.433                 & 0.293                 &1.55& 0.679 & 0.279     & 0.332    & \textbf{0.166}          & \textbf{0.128}        \\ \hline
\end{tabular}
\end{minipage}
\caption{{USB plug 3D pose estimation results. The metric reported here is the mean absolute error across the entire test set, with the unit in radians.}}
\label{tab:ablation_3d}
\end{table*}

\begin{table*}[t]
\centering
{
\begin{tabular}{||c|c||ccc||cccc||cc||c||}
\hline
 & \multirow{2}{*}{BYOL} & \multicolumn{3}{c||}{MAE} & \multicolumn{4}{c||}{T3}        & \multicolumn{2}{c||}{ResNet} & \multirow{2}{*}{UniT} \\\cline{3-11} 
          &                      & 0.25   & 0.5    & 0.75  & Tiny & Small & Medium & Large & Pretrain      & Random     &                       \\\hline
Rotation    & 1.202              & 0.310   & 0.235  & 0.284  & 1.551   & 0.914 & 0.306& 0.338  & 0.336     & 0.365        & \textbf{0.155}                  \\
Position & 11.2                  & 6.2  & 6.5  & 6.0  & 11.7 & 10.8 & 5.8 & 5.7        & 5.8          & 6.6      & \textbf{4.8} \\\hline                
\end{tabular}}
\caption{ {USB plug 6D pose estimation results. We report the mean absolute error for both rotation (radians) and position (millimeters) across the entire test set.}}
\label{tab:6D_benchmark}
\end{table*}

\begin{table*}[t]
\centering
{
\begin{tabular}{||c|c||ccc||cccc||cc||c||}
\hline
 & \multirow{2}{*}{BYOL} & \multicolumn{3}{c||}{MAE} & \multicolumn{4}{c||}{T3}        & \multicolumn{2}{c||}{ResNet} & \multirow{2}{*}{UniT} \\\cline{3-11} 
          &                       & 0.25   & 0.5    & 0.75  & Tiny & Small & Medium & Large & Pretrain      & Random     &                       \\\hline
Freeze  (\%)  & 85.0                  & 77.1   & 77.1   & 81.0  & 0    & 35.4  & 82.3   & 79.2  & 85.4          & 72.1          & \textbf{92.1}                  \\
Fine-Tune (\%)& {97.5}                  & 85.4   & 81.3   & 89.6  & 89.6 & 87.7  & 85.4   & 85.0  & \textbf{97.9}          & 85.4       & {97.3} \\\hline                
\end{tabular}}
\caption{{Classification benchmarks on YCB-Sight dataset \cite{suresh2021efficient}. The representations for BYOL, MAE, and UniT are all obtained using training data from all six objects.}  }
\label{tab:clf_benchmark}
\end{table*}

\begin{table*}[t]
\centering
{
\begin{tabular}{|c|c|c|c|c|c|c|c|}
\hline
          & all  & master chef can & sugar box & tomato soup can & potted meat can & bleach cleanser & wood block \\ \hline
Freeze  (\%)   & 92.1 & 87.9            & 90.0      & 86.3            & 87.9            & 91.0            & 89.0       \\ 
Fine-Tune(\%)  & 97.3 & 90.4            & 88.5      & 87.7            & 87.7            & 93.5            & 92.5       \\ \hline
\end{tabular}}
\caption{ {UniT representation training ablation study. ``all” indicates the representation trained on all six objects, while other columns, such as ``sugar box,” refer to representations trained only on data from that specific object. }}
\label{tab:clf_ablation}
\end{table*}

\begin{figure*}[t]
\centering
\begin{overpic}[trim=0 0 0 0,clip, width=0.90\textwidth]{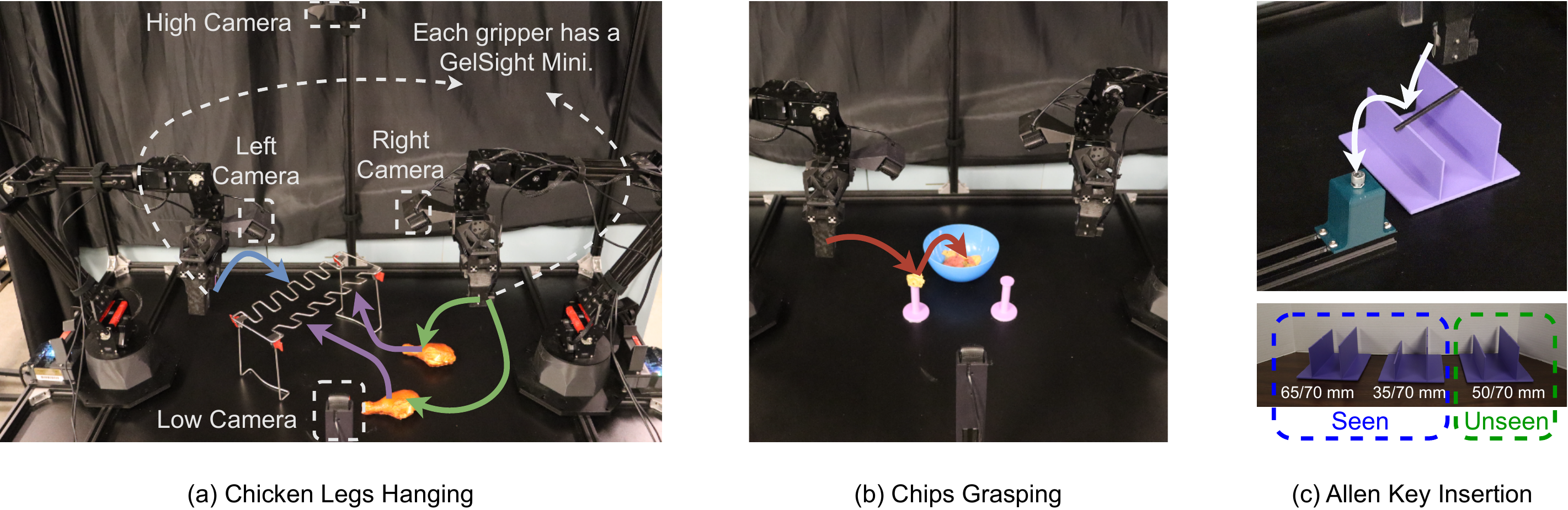}
\end{overpic}
\caption{Overview of three tasks and hardware setup. In (c), the heights on the left and right sides of the rack are 65/70 mm, 35/70 mm, and 50/70 mm. These varying heights result in different in-hand poses when the Allen key is grasped. The training dataset was collected using the 65/70 mm and 35/70 mm heights.}
\label{fig:setup}
\end{figure*}

\begin{table*}[t]
\centering
\begin{tabular}{|c||cc||c||cccc||}
\hline
                            & \multicolumn{2}{c||}{Chicken Legs Hanging} & \multirow{2}{*}{Chips Grasping} & \multicolumn{4}{c||}{Allen Key Insertion}                                                               \\ \cline{1-3} \cline{5-8} 
                            & \multicolumn{1}{c|}{One Leg Inserted} & Two Legs Inserted &                                 & \multicolumn{1}{c|}{65/70 mm} & \multicolumn{1}{c|}{35/70 mm} & \multicolumn{1}{c|}{50/70 mm} & Total \\ \hline
Vision-Only                 & \multicolumn{1}{c}{9/15}    & 6/15     & 8/15                            & \multicolumn{1}{c}{\textbf{8/10}}     & \multicolumn{1}{c}{3/10}     & \multicolumn{1}{c}{4/10}     & 15/30 \\ 
Visual-Tactile from Scratch & \multicolumn{1}{c}{8/15}    & 5/15     & \textbf{14/15}                           & \multicolumn{1}{c}{6/10}     & \multicolumn{1}{c}{5/10}     & \multicolumn{1}{c}{6/10}     & 17/30 \\
Visual-Tactile with UniT    & \multicolumn{1}{c}{\textbf{13/15}}   & \textbf{9/15}     & \textbf{14/15}                           & \multicolumn{1}{c}{\textbf{8/10}}     & \multicolumn{1}{c}{\textbf{7/10}}     & \multicolumn{1}{c}{\textbf{8/10}}     & \textbf{23/30} \\ \hline
\end{tabular}
\caption{Success rates of policy rollout on three different real tasks.}
\label{tab:policy_result}
\end{table*}

\begin{table*}[t]
\centering
{
\begin{tabular}{|c|c|c|c|c|c|c|}
\hline
             & Vision Only & Visual Tactile from Scratch & T3 Freeze & T3 Fine-Tune & UniT Freeze & UniT Fine-Tune \\ \hline
Success Rate & 0.501        & 0.502                        & 0.509      & 0.516         & \textbf{0.571}        & \textbf{0.575}           \\ \hline
\end{tabular}}
\caption{{Success rate of policy rollout on the simulated peg insertion task shown in Fig.~\ref{fig:task_tacsl}}.}
\label{tab:tacsl_result}
\end{table*}

\begin{figure}[t]
\centering
\begin{overpic}[trim=0 0 0 0,clip, width=0.48\textwidth]{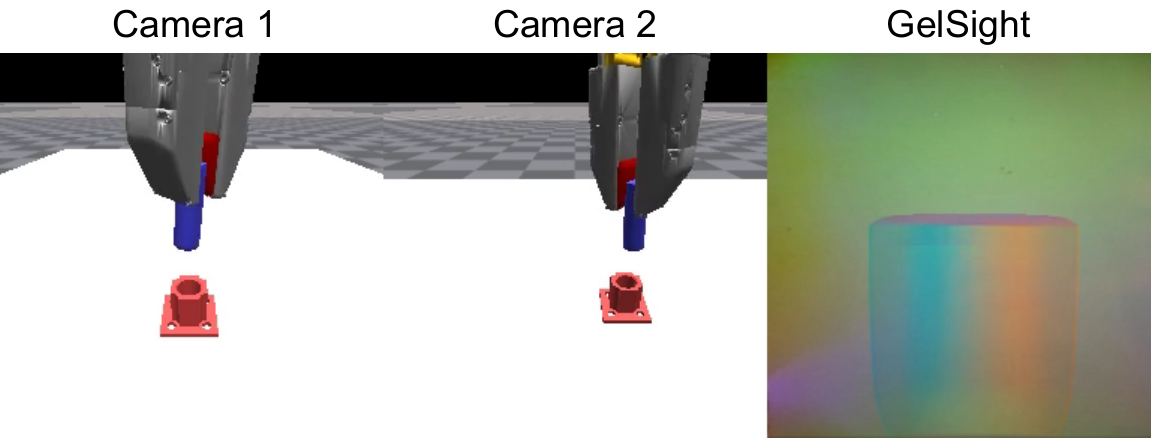}
\end{overpic}
\caption{{Simulated peg insertion task using TacSL \cite{akinola2024tacsl}.}}
\label{fig:task_tacsl}
\end{figure}

{In this section, the effectiveness of UniT is demonstrated on USB plug 3D pose and 6D pose estimation tasks, and tactile classification task.} We benchmark UniT with training a ResNet \cite{he2016deep} from scratch, as well as other representation learning frameworks: BYOL \cite{grill2020bootstrap}, MAE \cite{he2022masked}, and the state-of-the-art tactile representation framework, T3 \cite{zhao2024transferable}.  {In all experiments, we evaluate performance on the test set every 10 training epochs and report the average metric values from the last ten checkpoints.}

\subsection{Pose Estimation}
{The pose estimation tasks involve estimating the 3D or 6D pose of a USB plug based on its tactile image. Here, we conduct two types of experiments: in the 3D pose task, we estimate only the rotation, and the model output is a 4-dimensional quaternion. In the 6D pose task, we estimate both position and rotation simultaneously, with the model outputting a 7-dimensional vector (3-dimensional position + 4-dimensional quaternion). As illustrated in our supplementary video, changes in the 6D pose of the USB plug result in altered contact configurations and corresponding variations in the tactile image.} This type of pose estimation is crucial for robotic insertion tasks that require high precision. This task is challenging because minor differences in the tactile images can correspond to significantly different poses. The pose estimation model must accurately map these subtle tactile features to the correct pose.

{As shown in the supplementary video, we use the OptiTrack motion capture system to label the 6D pose ground truth. In the data collection process, human move the USB plug randomly and we record the raw tactile image as well as 6D pose ground truth in 10 Hz.} 

Directly splitting the training and test sets from continuous data can cause data leakage issues. To avoid this, we have collected data in the form of episodes. An episode consists of a sequence of continuous recorded data, with no continuity overlap between different episodes. Within each episode, the movement of the USB plug by humans is random, and there is no fixed paradigm across episodes. When dividing the training and test sets, we do not use individual images as the smallest unit; instead, we use entire episodes. As a result, the data in the test set is completely absent from the training set. We use seed 42 and maintain a 9:1 training-to-test ratio to split the episodes. Ultimately, we obtained 10,111 tactile images for training and 1,229 tactile images for testing. 

We use the following loss function to train the model, which converts quaternions to angle differences
$
    L = \mathbb{E} \left[ 2 \cdot \text{arccos}\left( \left| \hat{q} \cdot q \right| \right) \right],
$
where $\hat{q}$ and $q$ are estimated and z quaternions. {The loss function for 6D pose estimation includes quaternion-based angular error loss, supplemented with an additional MSE error loss for linear position. For more details, please refer to our open-source repository https://github.com/ZhengtongXu/UniT.}

{The benchmark results of 3D and 6D pose estimation tasks are shown in Tables~\ref{tab:ablation_3d} and \ref{tab:6D_benchmark}. } {A. K. represents Allen key dataset and S. B. represents small ball dataset. For 6D pose estimation, we test representations trained with Allen key data. For T3, we utilize the pretrained model released by the authors of \cite{zhao2024transferable}. We freeze the encoders of BYOL, T3, MAE, and UniT and train only the decoders on the pose estimation datasets. In the ResNet experiments, we train both the encoder and the decoders on pose data directly, as both ResNet and ResNet with pretrain do not go through a tactile representation learning process. The MAE with ViT-Base backbone has a model size of 143.20 M, whereas UniT with a representation dimension of $16\times20$ has a model size of 79.81 M.} Our experimental results clearly show that UniT delivered the best performance on both 3D pose and 6D pose estimation tasks.

{Finally, the ablation study for UniT in Table~\ref{tab:ablation_3d} demonstrates the contributions of VQ and discriminator to learning effective tactile representations. Removing VQ results in worse performance across all tested representation dimensions and for both the Allen key and small ball representations. For discriminator, except for the small ball with an $8\times10$ representation, most cases show better results when discriminator is included. Even in the small ball with $8\times10$ representation case, the performance with and without discriminator is quite similar. Overall, including discriminator generally enhances model performance.}

\subsection{Classification}

{Our classification experiment uses the real-environment tactile data from the YCB-Sight dataset \cite{suresh2021efficient}. The authors of YCB-Sight have released a visuo-tactile dataset including both simulation and real data. The real environment data includes data from six objects, each with 40 tactile images with contact with the object. We used the 240 contact images to build our classification dataset, labeling each image by the object it contacted. We shuffled these 240 images and split them into training and test sets in an 8:2 ratio. The benchmarking results are shown in Table~\ref{tab:clf_benchmark}. {We report classification accuracy on the test set as our evaluation metric. ``Freeze” means freezing the encoder and using the representation in a zero-shot manner to train the decoder on the classification training set. ``Fine-Tune” means training both the decoder and fine-tuning the encoder on the classification training set.} We tested two setups: one where the representation was transferred to this task in a zero-shot manner, and another where the encoders were fine-tuned specifically for this task. Our experimental results clearly show that UniT delivered the best performance in the zero-shot setup. When fine-tuned, UniT's performance was comparable to that of ResNet with pretraining and BYOL. The strong performance of ResNet with pretraining and BYOL in the fine-tuning setup can be attributed to the small size of the dataset, as BYOL’s backbone is also based on ResNet. On the other hand, for pose estimation, ResNet's performance is not as strong as UniT's, as shown in Tables \ref{tab:ablation_3d} and \ref{tab:6D_benchmark}. }

{Moreover, Table \ref{tab:clf_ablation} shows that, for UniT, a well-performing representation can be achieved with only a small amount of single-object data, allowing the trained representation to generalize in a zero-shot manner to unseen objects. This is similar to our use of the Allen key and small ball in pose estimation. By comparing Tables \ref{tab:clf_benchmark} and \ref{tab:clf_ablation}, UniT representations trained on one object like the sugar box or bleach cleanser achieve better zero-shot classification results than MAE and T3, even after fine-tuning. Notably, MAE representations here are trained on all six objects and T3 are pretrained on a large amount of tactile images.}

\section{Supervised Policy Learning Experiments}\label{sec:policy_exp}

In this section, the effectiveness of UniT in imitation learning is demonstrated. We use Aloha \cite{zhao2023learning} as our hardware platform. The hardware setup is shown in Fig.~\ref{fig:setup}(a). We integrate UniT into diffusion policy \cite{chi2023diffusionpolicy}, and benchmark with vision-only diffusion policy and visual-tactile diffusion policy with tactile encoders trained from scratch. We freeze the UniT encoder when integrating it into diffusion policy, as shown in Fig.~\ref{fig:decoder}. {All three methods are trained on the same dataset, with each method using a diffusion policy
as the main network, differing only in the structure of the tactile encoder, while the rest of the network architecture and hyperparameters are identical.} We completed three tasks, as shown in Fig.~\ref{fig:setup}. {The ``Chicken Legs Hanging", ``Chips Grasping", and ``Allen Key Insertion" datasets contain 200, 180, and 170 demonstrations, respectively. The action space is defined in terms of joint angles: each arm of the ALOHA dual-arm robot has 6 degrees of freedom, plus an additional degree of freedom for grasping width, totaling 7 degrees of freedom per arm. For the ``Allen Key Insertion" task, which involves only one arm, the policy outputs 7-DoF actions. In contrast, both the ``Chips Grasping" and `Chicken Legs Hanging" tasks are bimanual manipulation, so the policy outputs 14-DoF actions. The frequency of all policy rollouts is 10 Hz and the amount of inference denoising steps is 24.} More details of each task are as follows.

\textbf{Chicken Legs Hanging}: The chicken legs hanging task is a dual-arm manipulation task in which one robotic arm grasps the rack to prevent it from moving due to the pushing force exerted while hanging chicken legs. The other arm picks up two artificial chicken legs from a table and hangs them on the rack. {The rack slot has a trapezoidal shape, with a width ranging from 15.6 mm on the innermost side to 20.8 mm on the outermost side. In comparison, the chicken drumstick root varies in width from 19.2 mm at its narrowest to 22.2 mm at its widest. This close dimensional match necessitates precise alignment to securely hang the drumstick, underscoring the high level of precision required for successful task execution.} Moreover, this task involves rich force interactions with the objects being manipulated: the left arm applies force on the rack and stabilizes it, while the right arm exerts pressure to insert the chicken leg into the slot on the rack. In this situation, real-time tactile feedback is crucial for fully perceiving these interactions.

\textbf{Chips Grasping}: The Chips grasping task is also a dual-arm manipulation task. During this task, depending on whether the chips are on the left or right support stand, the corresponding robot arm picks up the chips and places them into a bowl. To accommodate the maximum width of the grippers, we resized the chips from their original dimensions. The challenge of this task lies in the fragility of the chips, which necessitates precise control over the gripper width. The gripper must adjust accurately to accommodate the shape and size of the chips. Inadequate adjustment may lead to either missing the grasp or crushing the chips. In this situation, real-time tactile feedback is crucial because relying solely on visual feedback makes it difficult to accurately determine the state of the grasp. This can easily lead to missing the grasp or breaking the chips when handling different sizes of chips.

\textbf{Allen Key Insertion}: The Allen key insertion task requires the robot to grasp a Allen key on a rack and then insert the Allen key into a nut. {We used a 6 mm nut and an Allen key with 5.5 mm head. Therefore, the precision requirement for this insertion is at the millimeter level. In this configuration, the Allen key can still be used to tighten and loosen the nut effectively after insertion. } Given that Aloha is a low-cost robot, its repeatability in terms of precision is not high, which adds significant challenges to this task.

{During data collection and policy rollout, we adjusted the height difference between the rack's supports, altering the Allen key's in-hand angle when grasped by the robot. Data was collected for two rack types in the training set, while policy rollout tested three types, including one entirely unseen rack representing out-of-distribution data, as shown in Fig.~\ref{fig:setup}(c).}

{The tactile modality plays a crucial role in this task: (1) It provides information on the Allen key's in-hand pose, enabling precise end-effector adjustments across varying rack types during insertion. (2) When initial alignments fail, tactile feedback is essential for interactively refining the policy to guide the Allen key into the nut accurately.}

The results of the policy rollout on these three tasks are shown in Table~\ref{tab:policy_result}. It is evident that the visual-tactile policy with UniT delivers the best performance on each task. As described earlier, the three tasks presented here involve rich robot-object-environment interactions. Our experimental results highlight two important points regarding these types of tasks: 1) The tactile modality is crucial for manipulation tasks with rich interactions, where relying solely on vision often fails to deliver optimal performance. 2) Compared to treating tactile images as regular visual images and training encoders from scratch, using UniT significantly enhances the effectiveness of integrating the tactile modality into the policy. For more detailed information about our experimental results, see the supplementary video and the project website {https://zhengtongxu.github.io/unit-website/}.

Besides experiments in real environment, we constructed a simulated peg insertion task using TacSL \cite{akinola2024tacsl}, as shown in Fig.~\ref{fig:task_tacsl}. To collect training data, we gathered 30 episodes through human demonstration with SpaceMouse. As shown in Fig.~\ref{fig:task_tacsl}, the observation provides two camera images from different viewpoints and one tactile image from GelSight. The action space consists of six-degree-of-freedom end-effector velocity control. During both data collection and policy rollout, the robot’s end effector is initialized in different configurations, and the hole position is randomized within a -5~cm to~5~cm range along the x and y-axes. During rollout, task success is determined by checking the spatial relationship between the peg and the socket. 

{We train all policies using the same learning rate scheduler for 500 epochs. Every 25 epochs, we perform rollouts across 50 different environment initializations to measure the success rate. The final metric is computed as the average success rate over the rollouts from the last 10 checkpoints. Moreover, we repeat the evaluation across three different seeds and report the average results in Table~\ref{tab:tacsl_result}. Our results demonstrate that policies with UniT outperform those with T3, as well as vision-only policies and visual-tactile policies trained from scratch.}

\section{Future Work}

For future work, we believe data-efficient tactile representation learning can be further improved by incorporating a diverse set of objects with fine textures to enhance generalization and performance. Specifically, investigating how training with a broader range of physical objects, including those with intricate fabric-like patterns, can enrich the representation’s ability to capture subtle tactile details is a promising direction.

\ifCLASSOPTIONcaptionsoff
  \newpage
\fi

% ==========   Bibliography
\bibliographystyle{IEEEtran}
\bibliography{IEEEabrv,paperref}

\end{document}